\documentclass[runningheads]{llncs}
\usepackage{makeidx}
\usepackage{graphicx}
\usepackage{amsmath,amssymb} 
\usepackage{color}
\usepackage{mathtools}
\usepackage{multicol}
\usepackage{subcaption}
\usepackage{wrapfig}
\usepackage{booktabs}
\usepackage{bbm}
\usepackage{icomma}

\begin{document}
	\pagestyle{headings}
	\mainmatter

	\def\GCPR20SubNumber{30}

	\title{Neural Architecture Performance Prediction Using Graph Neural Networks }
    \titlerunning{Neural Architecture Performance Prediction Using Graph Neural Networks}
	\authorrunning{J.Lukasik, D.Friede, H.Stuckenschmidt, M.Keuper}
	\author{Jovita~Lukasik \orcidID{0000-0003-4243-9188}, David~Friede, Heiner~Stuckenschmidt \orcidID{0000-0002-0209-3859}, Margret~Keuper \orcidID{0000-0002-8437-7993}}
	\institute{University of Mannheim, Germany \\
	\texttt{\{jovita, david, heiner\}@informatik.uni-mannheim.de, keuper@uni-mannheim.de} }

	\maketitle

	\begin{abstract}
		In computer vision research, the process of automating architecture engineering, Neural Architecture Search (NAS), has gained substantial interest. Due to the high computational costs, most recent approaches to NAS as well as the few available benchmarks only provide limited search spaces. In this paper we propose a surrogate model for neural architecture performance prediction built upon Graph Neural Networks (GNN).  We demonstrate the effectiveness of this surrogate model on neural architecture performance prediction for structurally unknown architectures (i.e. zero shot prediction) by evaluating the GNN on several experiments on the NAS-Bench-101 dataset.
	\end{abstract}

    \section{Introduction}
    Deep learning using convolutional neural architectures has been the driving force of recent progress in computer vision and related domains. Multiple interdependent aspects such as the increasing availability of training data and compute resources are responsible for this success. Arguably, none has had as much impact as the advancement of novel neural architectures \cite{krizhevsky2012imagenet,goodfellow2014generative}. Thus, the focus of computer vision research has shifted from a feature engineering process to an architecture engineering process. The direct consequence is the need to automate this process using machine learning techniques.

\textit{Neural Architecture Search} (NAS) \cite{elsken2018neural} attends to techniques automating architecture engineering. Due to very long compute times for the recurrent search and evaluation of new candidate architectures~\cite{zoph2018learning}
, NAS research has hardly been accessible for researchers without access to large-scale compute systems.
Yet, the publication of \textit{NAS-Bench-101} \cite{ying2019bench}, a dataset of over $423$k fully trained neural architectures, facilitates a paradigm change in NAS research. Instead of carefully evaluating each new proposed neural architecture, NAS-Bench-101 enables to experiment with classical data-based methods such as supervised learning to evaluate neural architectures. While the impact of benchmarks such as NAS-Bench-101 on the community is high, they come at extreme computational costs. All architectures in the search space have to be extensively evaluated, which leads to practical restrictions on the search space. This calls for accurate surrogate models that enable to extrapolate expected performances to structurally different and larger architectures in unseen areas of the search space, i.e.\ \textit{zero shot prediction}. 
In this paper, we first tackle the task of learning to predict the accuracy of convolutional neural architectures in a supervised way, i.e. we learn a surrogate model that enables to predict the performance of neural architectures on the CIFAR-10 image classification task. Furthermore, we evaluate our proposed model on two different zero shot prediction scenarios and show its ability to accurately predict performances in previously unseen regions of the search space. 

Most current neural architectures for computer vision can be represented as directed, acyclic graphs (DAGs). Thus, we base our surrogate model on Graph Neural Networks. \textit{Graph Neural Networks} (GNNs) \cite{wu2019comprehensive} have proven to be very powerful comprehending local node features and graph substructures. This makes them a very useful tool to embed nodes as well as full graphs like the NAS-Bench-101 architectures into continuous spaces.
Furthermore, the benefit of GNNs over Recurrent Neural Networks (RNNs) has been shown in the context of graph generating models.
The model \textit{Deep Generative Models of Graphs} (DGMG) in \cite{li2018learning} utilizes GNNs and shows dominance over RNN methods. 
DGMG is able to capture the structure of graph data and its attributes in a way that probabilistic dependencies within graph nodes and edges can be expressed, yielding in learning a distribution over any graph. This makes DGMG a strong tool to map neural architectures into a feature representation which captures the complex relation within the neural architecture. 

Inspired by \cite{li2018learning}, we utilize the GNN as our surrogate model for the performance prediction task.

In summary, in this paper we make the following contributions: We present a surrogate model-- a graph encoder built on GNNs-- for neural architecture performance prediction trained and evaluated on the NAS-Bench-101 benchmark and show that this neural performance predictor accurately predicts architecture performances in previously structurally different and unseen regions of the search space, i.e. zero shot prediction.

The remaining paper is structured as follows: Section \ref{sec:related} gives a short review of the related work.  In Section \ref{sec:encoder} we present our proposed encoder model. Section \ref{implementation_details} gives detailed model implementation details of the proposed surrogate model. In Section \ref{sec:between}, we describe the NAS-Bench-101 dataset on which we conduct our evaluation. In Section \ref{sec:experiments}, we present our experiments and results. Finally, we give a conclusion and outline some future directions in Section \ref{sec:conclusion}.

    \section{Related Work} \label{sec:related}
    \subsection{Neural Architecture Search}

Neural Architecture Search (NAS) \cite{Zoph2017,Real2017,zoph2018learning,Liu2018,Pham2018}, the process of designing neural network architectures in an automatic way, gained substantial attention recently. See \cite{elsken2018neural} for an overview and detailed survey over recent NAS methods. The currently most successful approaches follow different paradigms: Reinforcement learning (RL) \cite{Zoph2017,zoph2018learning,Pham2018} as a NAS strategy considers the neural architecture generation as the agent's action with it's reward given in terms of validation accuracy. Evolutionary Algorithm (EA) \cite{Real2017,Liu2017} approaches optimizing the neural architectures themselves by guiding the mutation of architectures and evaluating their fitness given in terms of validation accuracy. Bayesian optimization (BO) \cite{Kandasamy2018} derives kernels for architecture similarity measurements to extrapolate the search space. Gradient based methods \cite{Liu2018,luo2018neural} use continuous relaxations of neural architectures to allow for gradient-based optimization.

\subsection{Neural Architecture Benchmark Datasets}
NAS-Bench-101 \cite{ying2019bench} is a public dataset of 423k neural architectures and provides tabular benchmark results for a restricted cell structured architecture search space \cite{zoph2018learning} with exhaustive evaluation on the CIFAR-10 image classification dataset \cite{Krizhevsky2012}. As shown in \cite{zela2020nasbenchshot}, only subspaces of the architectures in NAS-Bench-101 can be used to evaluate one-shot NAS methods~\cite{Liu2018,Pham2018}, motivating their proposed variant NAS-Bench-1shot1~\cite{zela2020nasbenchshot}.

Similarly to NAS-Bench-101, NAS-Bench-201~\cite{dong20}
uses a restricted, 
cell- structured search space, 
while the employed graph representation allows to evaluate discrete and one-shot NAS algorithms. The search space is even more restricted than NAS-Bench-101, providing only 6k unique evaluated architectures in total. We conduct our experiments on NAS-Bench-101, which is the largest available tabular neural architecture benchmark for computer vision problems.

\nocite{white2019bananas}

\subsection{Performance Predictor for Neural Architectures}
The work on performance prediction models for neural architectures is very limited. \cite{PNAS} uses a performance predictor in an iterative manner during the search process of NAS. \cite{accperf} uses features of a neural architecture, such as the validation accuracy, some architecture parameters such as the number of weights and the number of layers as well as hyperparameters, to predict learning curves during the training process by means of a SRM regressor. \cite{luo2018neural} proposes a performance prediction model learned in combination with an auto-encoder in an end-to-end manner. The neural architectures are mapped into a latent feature representation, which is then used by the predictor for performance prediction and are further decoded into new neural architectures.
Recently \cite{semisupervised} proposes a semi-supervised assessor of neural architectures. The graphs are employed by an auto-encoder to discover latent feature representations, which is then fine-tuned by means of a graph similarity measurement. Lastly, a graph convolution network is used for performance prediction.

\subsection{Graph Neural Networks}
Combining modern machine learning methods with graph structured data has increasingly gaining popularity. 
One can interpret it as an extension of deep learning techniques to non-Euclidean data \cite{bronstein2017geometric} or even as inducing relational biases within deep learning architectures to enable combinatorial generalization \cite{battaglia2018relational}. 
Because of the discrete nature of graphs, they can not trivially be optimized in differentiable learning methods that act on continuous spaces. The concept of graph neural networks is a remedy to this limitation.
The idea of Graph Neural Networks as an iterative process which propagates the node states until an equilibrium is reached,  was initially mentioned in 2005 \cite{gori2005new}. Motivated by the increasing populatiry of CNNs, \cite{bruna2013spectral} and \cite{henaff2015deep} defined graph convolutions in the Fourier domain by utilizing the graph Laplacian. The modern interpretation of GNNs was first mentioned in \cite{gated,niepert2016learning,kipf2016semi} where node information was inductively updated through aggregating information of each node's neighborhood. This approach was further specified and generalized by \cite{hamilton2017inductive} and~\cite{gilmer2017neural}.

The research in GNNs enabled breakthroughs in multiple areas related to graph analysis such as computer vision \cite{xu2017scene,landrieu2018large,yi2017syncspeccnn}, natural language processing \cite{bastings2017graph}, recommender systems \cite{monti2017geometric}, chemistry \cite{gilmer2017neural} and others. The capability of GNNs to accurately model dependencies between nodes makes them the foundation of our research. We utilize them to move from the discrete graph space to the continuous space.

In this paper, we want to use continuous methods, GNNs, to handle the graphs characterizing neural architectures from the NAS-Bench-101 dataset~\cite{ying2019bench}. More precisely, we show the ability of GNNs to encode neural architectures such as to allow for a regression of their expected performance on an image classification problem.

    \color{black}
    \section{The Graph Encoder}\label{sec:encoder}
    \begin{figure}[t]
\begin{center}
\includegraphics[height=4.2cm]{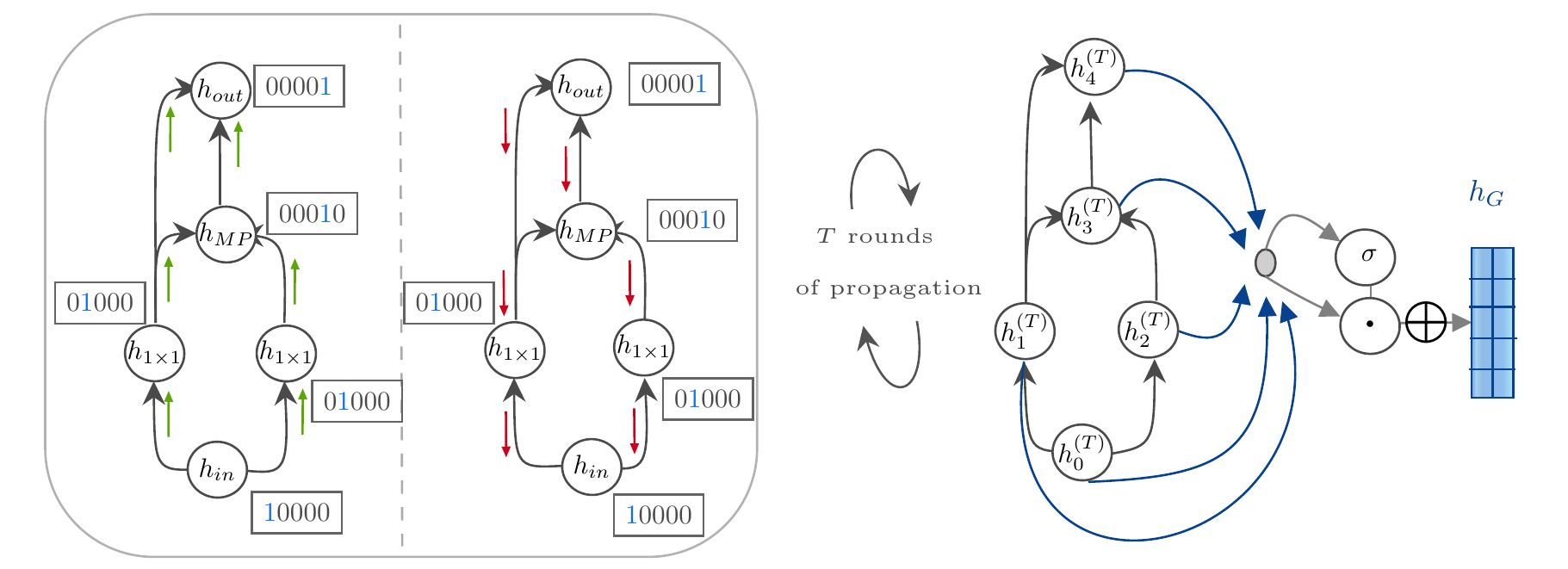}
\end{center}
   \caption{Illustration of the graph encoding process: The node-level propagation using $T$ rounds of bidirectional message passing (\textit{left}) and the graph-level aggregation into a single graph embedding $h_G$ (\textit{right}).}
\label{fig:encoder}
\end{figure}

In this section we present our GNN-based model to encode the discrete graph space of NAS-Bench-101 into a continuous vector space. One can imagine a single GNN iteration as a two-step procedure. First, each node sends out a message to its neighbors alongside its edges. Second, each node aggregates all incoming messages to update itself. After a final amount of these iteration steps, the individual node embeddings are aggregated into a single graph embedding.

\subsection{Node-Level Propagation} Let $G=(V,E)$ be a graph with nodes $v\in V$ and edges $e\in E\subseteq V\times V$. We denote $N(v)=\{u\in V\mid (u,v)\in E\}$ and $N^{out}(v)=\{u\in V\mid (v,u)\in E\}$ as the directed neighborhoods of a node $v \in V$. For each node $v\in V$, we associate an initial node embedding $h_v\in \mathbb{R}^{d_n}$. In our experiments we use a learnable look-up table based on the node types. Propagating information through the graph can be seen as an iterative \textit{message-passing} process

\begin{align}
m_{u\rightarrow v} &=  \Xi_{u\in N(v)}\bigl(M^{(t)}(h_v^{(t-1)}, h_u^{(t-1)})\bigr),\label{M module} \\ 
h_v^{(t)} &= U^{(t)}(h_v^{(t-1)}, m_{u\rightarrow v}), \label{U module}
\end{align}

with a differentiable message module $M^{(t)}$ in (\ref{M module}), a differentiable update module $U^{(t)}$ in (\ref{U module}) and a differentiable, permutation invariant aggregation function $\Xi$. The message module $M^{(t)}$ is illustrated by the green arrows in Figure \ref{fig:encoder} (left). To address the directed nature of the NAS-Bench-101 graphs, we add a reverse message module

\begin{align}
m^{out}_{u\rightarrow v} &= \Xi_{u\in N^{out}(v)}\bigl(M_{out}^{(t)}(h_v^{(t-1)}, h_u^{(t-1)}\bigr),\label{M rev}\\
h_v^{(t)} &= U^{(t)}(h_v^{(t-1)}, m_{u\rightarrow v}, m^{out}_{u\rightarrow v}).
\end{align}

This is outlined in Figure \ref{fig:encoder} (left) by the red arrows and leads to so-called bidirectional
message passing. The update module $U^{(t)}$ utilizes each node's incoming messages to update that node's embedding from $h_v^{(t-1)}$ to $h_v^{(t)}$.

Exploring many different choices for the message and update modules experimentally, we find that the settings similar to \cite{li2018learning} work best for our needs. We pick a concatenation together with a single linear layer for our message modules. The update module consists of a single gated recurrent unit (GRU) where $h_v^{(t-1)}$ is treated as the hidden state. For the aggregation function, we choose the sum. To increase the capacity of our model, on the one hand, we apply multiple rounds of propagation and on the other hand, we use a different set of parameters for each round.

\subsection{Graph-Level Aggregation} After the final round of message-passing, the propagated node embeddings $h=(h_v)_{v\in V}$ are aggregated into a single graph embedding $h_G\in \mathbb{R}^{d_g}$, where

  \begin{align}\label{graph aggr}
  	h_G = A(h),
  \end{align}

We obtain good results by using a linear layer combined with a gating layer that adjusts each node's fraction in the graph embedding. This aggregation layer $A$ in (\ref{graph aggr}) is further illustrated in Figure \ref{fig:encoder} (right).

    \section{Model Details}\label{implementation_details}
    In this section, we give further details on the implementation of our GNN model.

\subsection{Message}
The message module $M^{(t)}$ concatenates the embedding of the considered node $h_v^{(t-1)}$ as well as the incoming embedding $h_u^{(t-1)}$, each of dimension $d_n$. 
It further performs a linear transformation on the concatenated embedding. The reverse message module $M_{out}^{(t)}$ is a clone of $M^{(t)}$ initialized with its own weights,

\begin{align}
M^{(t)} &= \mathrm{Lin}_{2d_n \times 2d_n}\bigl([h_v^{(t-1)}, h_u^{(t-1)}]\bigr),\\
M_{out}^{(t)} &= \mathrm{Lin'}_{2d_n \times 2d_n}\bigl([h_v^{(t-1)}, h_u^{(t-1)}]\bigr).
\end{align}

The message module (green) and the reverse message (red) can be seen on the left side of Figure \ref{fig:encoder}.

\subsection{Update}
The update module $U^{(t)}$ is a single GRU cell. First, the incoming messages $m_{u\rightarrow v}$ and $m^{out}_{u\rightarrow v}$ are added and handled as the GRU input. Second, the node embedding $h_v^{(t-1)}$ is treated as the hidden state and is updated,

\begin{align}
U^{(t)} = \mathrm{GRUCell}_{2d_n, d_n}\bigl(m_{u\rightarrow v} + m^{out}_{u\rightarrow v},\  h_v^{(t-1)}\bigr).
\end{align}

\subsection{Aggregation}
We use two rounds of propagation before aggregating the node embeddings into a single graph embedding. This graph aggregation consists of two parts. First, a linear layer transforms the node embeddings to the required graph embedding dimension $d_g$. Second, another linear layer combined with a sigmoid handles each node's fraction in the graph embedding,

\begin{align}
A_1 &= \mathrm{Lin}_{d_n \times d_g}(h_v^{(2)}),\\
A_2 &= \sigma\bigl(\mathrm{Lin}_{d_n \times 1}(h_v^{(2)})\bigr),\\
A &= \sum_v A_1 \odot A_2.
\end{align}
 
An illustration of the aggregation module is given in Figure \ref{fig:encoder} (right).

    \section{The NAS-Bench-101 Dataset}\label{sec:between}
NAS-Bench-101 \cite{ying2019bench} is a public dataset of neural architectures in a restricted cell structured search space \cite{zoph2018learning} evaluated on the CIFAR-10-classification set \cite{Krizhevsky2012}. NAS-Bench-101 considers the following constraints to limit the search space: it only considers directed acyclic graphs, the number of nodes is limited to $ \vert V\vert \leq 7$, the number of edges is limited to $\vert E\vert \leq 9$ and only 3 different operations are allowed $\{3~\times~3 ~\mathrm{convolution}, 1~\times~1 ~\mathrm{convolution}, 3~\times~3 ~\mathrm{max-pool} \}$. These restrictions lead to a total of $423$k unique convolutional architectures, which are built from the cells in the following way:
Each cell is stacked three times, followed by a max-pooling layer which reduces the feature map size by factor two. This pattern is repeated 3 times, followed by global average pooling and
a dense softmax layer to produce the output. While this search space is limited it covers relevant architectures such as for example ResNet like~\cite{resnet} and InceptionNet like~\cite{inceptionnet}  models~\cite{ying2019bench}. 

The architectures have been trained for four increasing numbers of epochs $\{4,12,36,108\}$. Each of these architectures is mapped to its test, validation and training measures. In this paper we use the architectures trained for $108$ epochs and aim to predict their corresponding validation and test accuracy.

    \section{Experiments}\label{sec:experiments}

We conduct experiments in three complementary domains. First, we evaluate the performance prediction ability of the proposed GNN in the traditional supervised setting. Then, we conduct zero shot prediction experiments in order to show the performance of the proposed model for unseen graph structures during training. Both experiments are carried out on the validation accuracies reported in NASBench101. Last, we compare our results to the recent publication by Tang et al.~\cite{semisupervised} in terms of test accuracy prediction.

\subsubsection{Implementation Details.} If not mentioned differently, we set $d_n=250$ for the node dimensions and $d_g=56$ for the dimension of the latent space. We split the dataset $70\%$/$20\%$/$10\%$ edit-sampled into training-, test- and validation set. 
All our experiments are implemented using PyTorch \cite{paszke2017automatic} and PyTorch Geometric~\cite{fey2019fast}. 

 The hidden layers of the regressor are of size $28, 14$ and $7$. We used no activation function for the very last output (linear regression) and trained the joint encoder model with a learning rate of~$1e^{-5}$ for 100 epochs. The hyperparameters were tuned with BOHB \cite{BOHB},

\subsection{Performance Prediction}\label{sec:pp}
\begin{figure}[t]
  \centering
    \includegraphics[height=3.05cm]{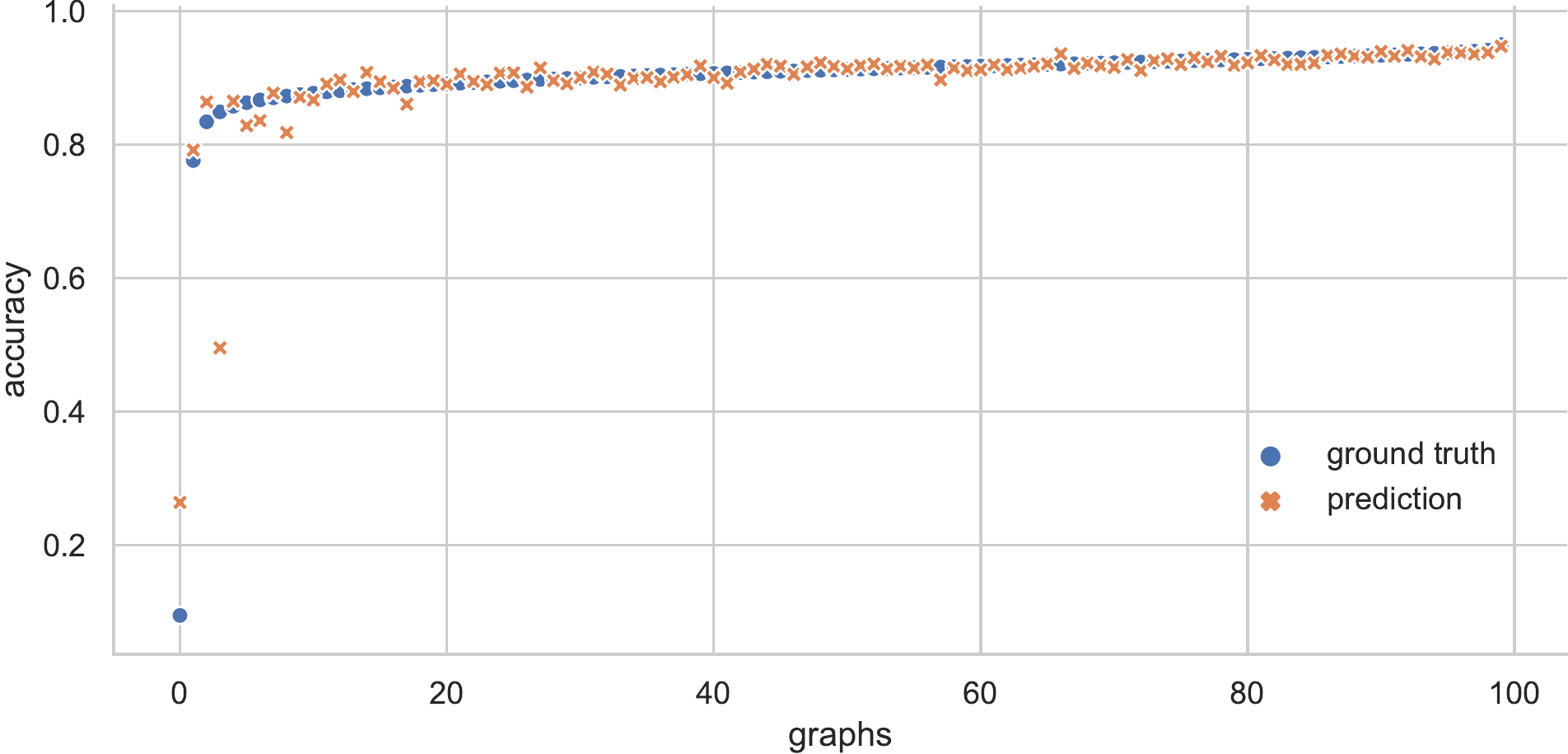}\hspace{1cm}
    \includegraphics[height=3.05cm]{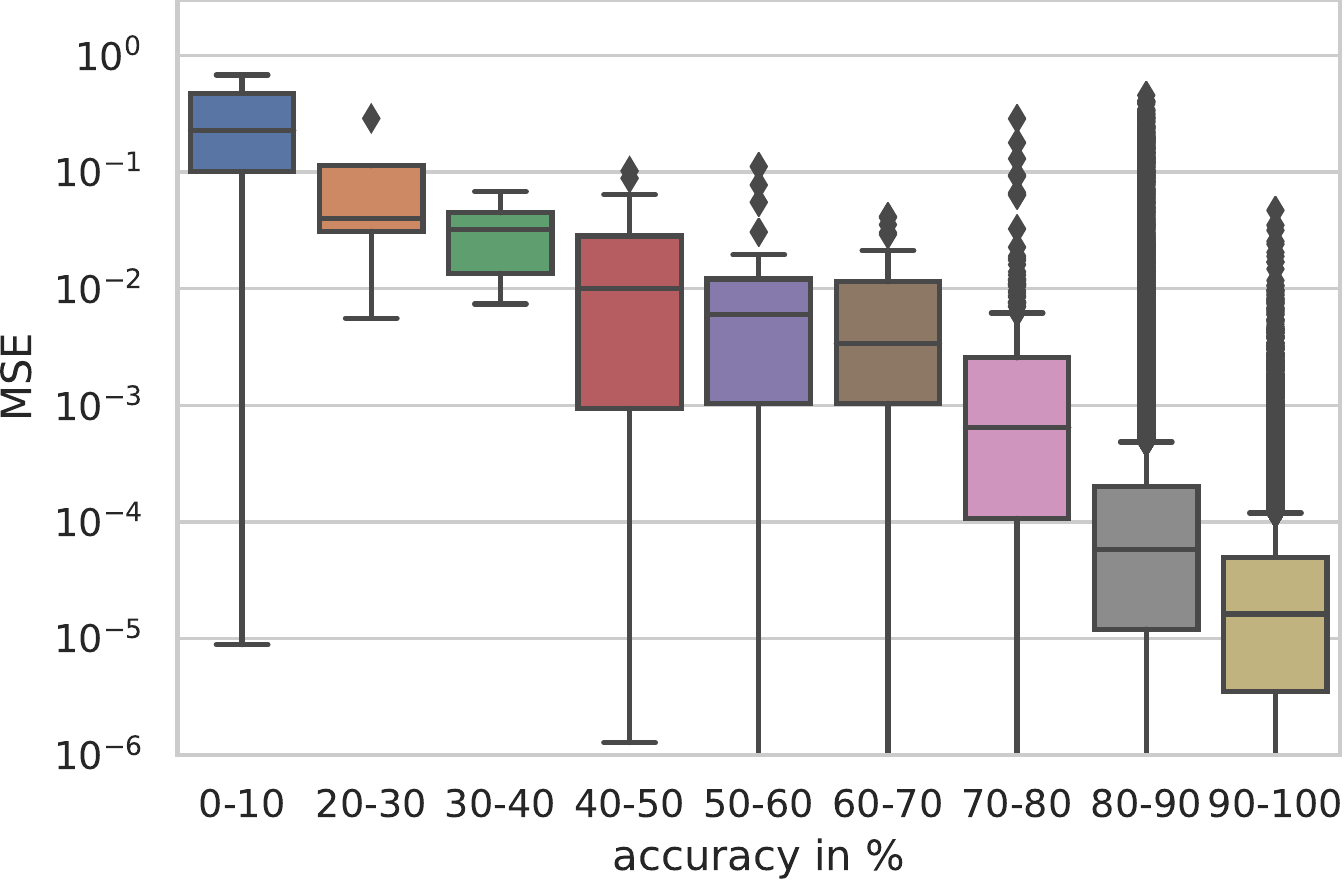}
  \caption{(\textit{Left})The predicted accuracy and ground truth of 100 randomly sampled graphs from the NAS-Bench-101 dataset showing a low prediction error for graphs with high accuracy. For low accuracy architectures, our model mostly predicts low values. (\textit{Right}) The mean and variance of the squared error of the test set performance prediction sorted by the ground truth accuracy in logarithmic scale. Predictions are very reliable for architectures in the high accuracy domain while errors are higher for very low performing architectures.} \label{fig:graphs}\label{fig:losses}
\end{figure}
\subsubsection{Supervised Performance Prediction.} Here, we evaluate the latent space generated by the encoder  with respect to its prediction error regarding a metric of interest of the NAS-Bench-101 graphs, i.e. the validation accuracy on CIFAR-10. For this purpose, we utilize a simple predictor, i.e. a four-layer MLP with ReLU non-linearities. 

We jointly train the encoder and the predictor supervisedly end-to-end. We test for prediction as well as for zero shot prediction errors. There are a few outliers in the NAS-Bench-101 graphs that end up with a low validation accuracy on the CIFAR-10 classification task. Figure \ref{fig:graphs} (left) visualizes these outliers and shows that our model is able to find them even if it cannot perfectly predict their accuracies. One can see that the model predicts the validation accuracy of well performing graphs very accurately. To further explore the loss, Figure \ref{fig:losses} (right) illustrates the mean and variance of the squared error of the test set partitioned in 9 bins with respect to the ground truth accuracy. The greater part of the loss arises from graphs with a low accuracy. More importantly, our model is very accurate in its prediction for graphs of interest namely graphs with high accuracy. 

\begin{figure}[t]
    \centering
        \includegraphics[height=2.53cm]{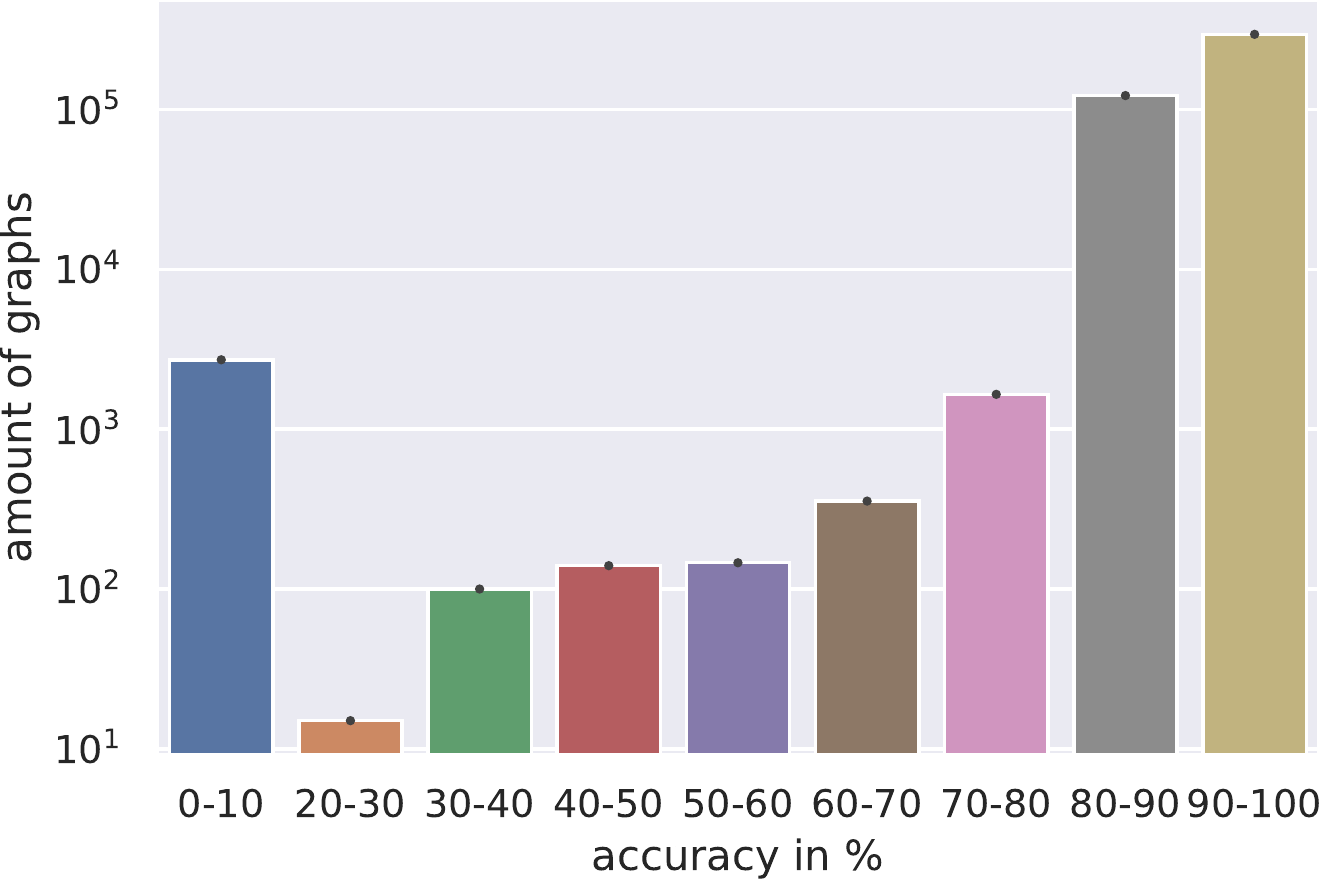}
        \includegraphics[height=2.53cm]{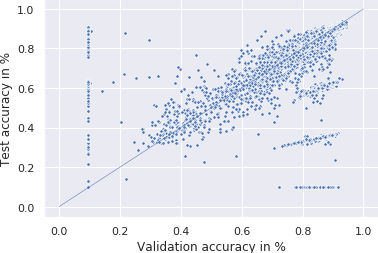}
        \includegraphics[height=2.53cm]{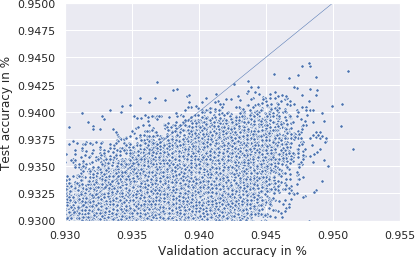}

    \caption{Distinct properties of NAS-Bench-101; The allocation of the dataset sorted by the ground truth accuracy in logarithmic scale  $\sim$98.8\% in the two last bins (\textit{left}). NAS-Bench-101 validation and test accuracy behaviour on the CIFAR-10 image classification task. Validation accuracy in $\%$ compared to the test accuracy in $\%$ of the neural architectures in the NAS-Bench-101 dataset (\textit{middle}). A more precise look into the areas of interest for neural architectures display that the best neural architecture by means of the test accuracy is unequal to the best accuracy by means of the validation accuracy (\textit{right}).}\label{fig:numbers}
\end{figure}
\begin{table}[t]
\caption{Predictive performance of the GNN encoder in terms of RMSE on supervised validation performance prediction.}
\centering
\begin{tabular}{lcc}
\toprule
Model                         & \multicolumn{1}{c}{Prediction}                         &  \\ \midrule
Encoder         & $\mathbf{0.0486 (\pm 0.1\%)}$   &  \\ \midrule
RF- Wide-Depth Feature Encoding \hspace{1cm}     & $0.061 (\pm 0.4\%)   $&  \\
RF- One-Hot Encoding         &$ 0.0632 (\pm 0.01\%)$ &  \\
MLP- One-Hot Encoding          &$ 0.0632 (\pm 0.02\%)$  & \\
RNN One-Hot Encoding      & $0.063 (\pm 0.01\%)   $&  \\
\bottomrule
\end{tabular}\label{tab:rmse}
\end{table}

The rather bad prediction of graphs with low and intermediate accuracy can be explained through their low share in the dataset. Taking a look at the distribution of the individual accuracies in the overall NAS-Bench-101 dataset, as shown in Figure \ref{fig:numbers} (left), illustrates the low share of low and intermediate accuracies in the dataset and explains therefore, the rather bad prediction behaviour of our surrogate model. Figure \ref{fig:numbers} (middle) and (right) plot the validation accuracy compared to the test accuracy of the NAS-Bench-101 dataset. This figure illustrates that predicting the best architecture on the validation set does not necessarily imply a proper prediction on the test set. 
 
We compare the results of the encoder 
to several baselines. Our baselines are a random forest approach and also an MLP regressor with four layers, 
using one-hot node feature encodings and graph depth/width feature encodings.
In order to compare to an RNN baseline, we adapted the RNN-surrogate model from \cite{PNAS}, which, in their original implementation, only handles cells of equal length. For the application to NAS-Bench-101 with cell types of different length, we do the following slight modification: 
we input to the LSTM a 0-padded one-hot vector of node attributes, encoding up to 7 nodes and 5 operations.

Table \ref{tab:rmse} summarises the performance prediction results on the supervised performance prediction task. All experiments are repeated $3$ times and we report the mean and the relative standard deviation. 
The experiments show that our surrogate model is able to predict the neural architecture performances in a stable way and outperforms all baselines in terms of the RMSE by a significant margin.

\begin{table}[t]
\caption{Predictive performance of the GNN encoder in terms of RMSE on the two different zero shot validation performance prediction tasks.}
\centering
\begin{tabular}{lccc}
\toprule
Model                                      & \multicolumn{2}{c}{zero shot Prediction}                               &  \\
                                                          & \footnotesize{$2, 3, 4, 5, 7 -6$}                     & \footnotesize{$2, 3, 4, 5, 6 -7$}                   &  \\ \midrule
Encoder           &$ \mathbf{0.0523(\pm 3.9 \%)}   $& $\mathbf{0.0573(\pm 1.7 \%)}$ &  \\ 
\midrule
RF- Wide-Depth Feature Encoding  \hspace{1cm}  & $0.06 (\pm 0.2\%) $& $0.073 (\pm 0.5\%) $&  \\
RF- One-Hot Encoding        &$ 0.07 (\pm 0.04\%)  $& $0.063 (\pm 0.1\%)$ &  \\
MLP- One-Hot Encoding     & $ 0.0647 (\pm 2.4\%)$ &$ 0.094 (\pm 12.7\%)$ &  \\
RNN One-Hot Encoding   & $0.062 (\pm 4.7\%) $& $0.069 (\pm 3.3\%) $& \\
\bottomrule
\end{tabular}\label{tab:zero_sho_rmse}
\end{table}
\subsubsection{Zero shot Performance Prediction.} Next, we consider the task of predicting the validation accuracy of structurally unknown graph types, i.e. zero shot prediction.
The zero shot prediction task is furthermore divided into two subtasks. First, the encoder is trained on all graphs of length ${2, 3, 4, 5, 7}$ and tested on graphs of length $6$. In this scenario, the unseen architectures could be understood as interpolations of seen architectures. Second, we learn the encoder on graphs of length ${2, 3, 4, 5, 6}$ and test it on graphs of length $7$. This case is expected to be harder not only because the graphs of length $7$ are the clear majority and have the highest diversity, but also because the prediction of their performance is an extrapolation out of the seen training distribution. 

Table \ref{tab:zero_sho_rmse} summarizes the performance prediction results on the zero shot performance prediction task. All experiments are repeated $3$ times and we report the mean and the relative standard deviation. As expected, the resulting RMSE is slightly higher for the extrapolation to graphs of length $7$ than for the zero shot prediction for graphs of length $6$. Yet the overall prediction improves over all baselines by a significant margin. The higher standard deviation in comparison to the random forest baselines indicates that the performance of the GNN depends more strongly on the weight initialization than in the fully supervised case. Yet, please note that this dependence on the initialization is still significantly lower than for the MLP and RNN baselines. The experiments show that our surrogate model is able to accurately predict data that it has never seen, i.e.~that it can predict the accuracies even for architectures not represented by the training distribution.

\subsection{Training Behaviour}
\begin{figure}[t]
  \centering
     \includegraphics[height=3.5cm]{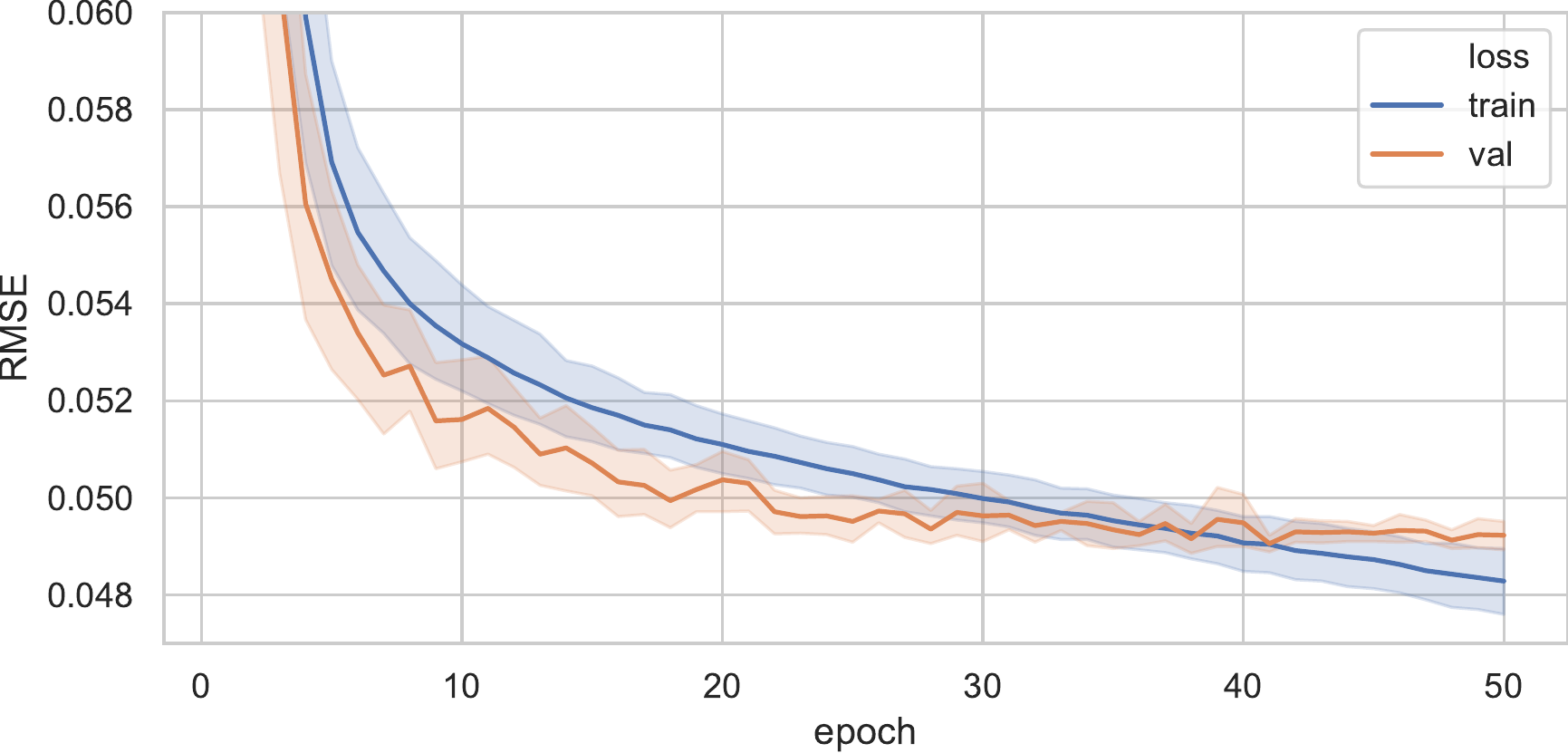} 
  \caption{Progress of loss and validation error over 50 epochs regarding performance prediction. Best validation RMSE $\sim0.0487$.} \label{fig:Interpolation}
 \end{figure}
In the following, we analyse the training behaviour of our model in the different scenarios described above.
\subsubsection{Supervised Performance Prediction.} For visualisation aspects of the training behaviour of our encoder, we plot the development of the training loss against the validation loss for the supervised performance prediction from Section $6.1$ \textit{Supervised Performance Prediction}. Figure \ref{fig:Interpolation} displays this development of training loss against validation loss measured by means of the RMSE. The smallest achieved RMSE is $\sim0.0487$ for training on $70\%$ of the dataset, i.e. $296,558$ samples. 

 \begin{figure}
     \begin{subfigure}[b]{0.5\textwidth}
         \includegraphics[width=0.95\textwidth]{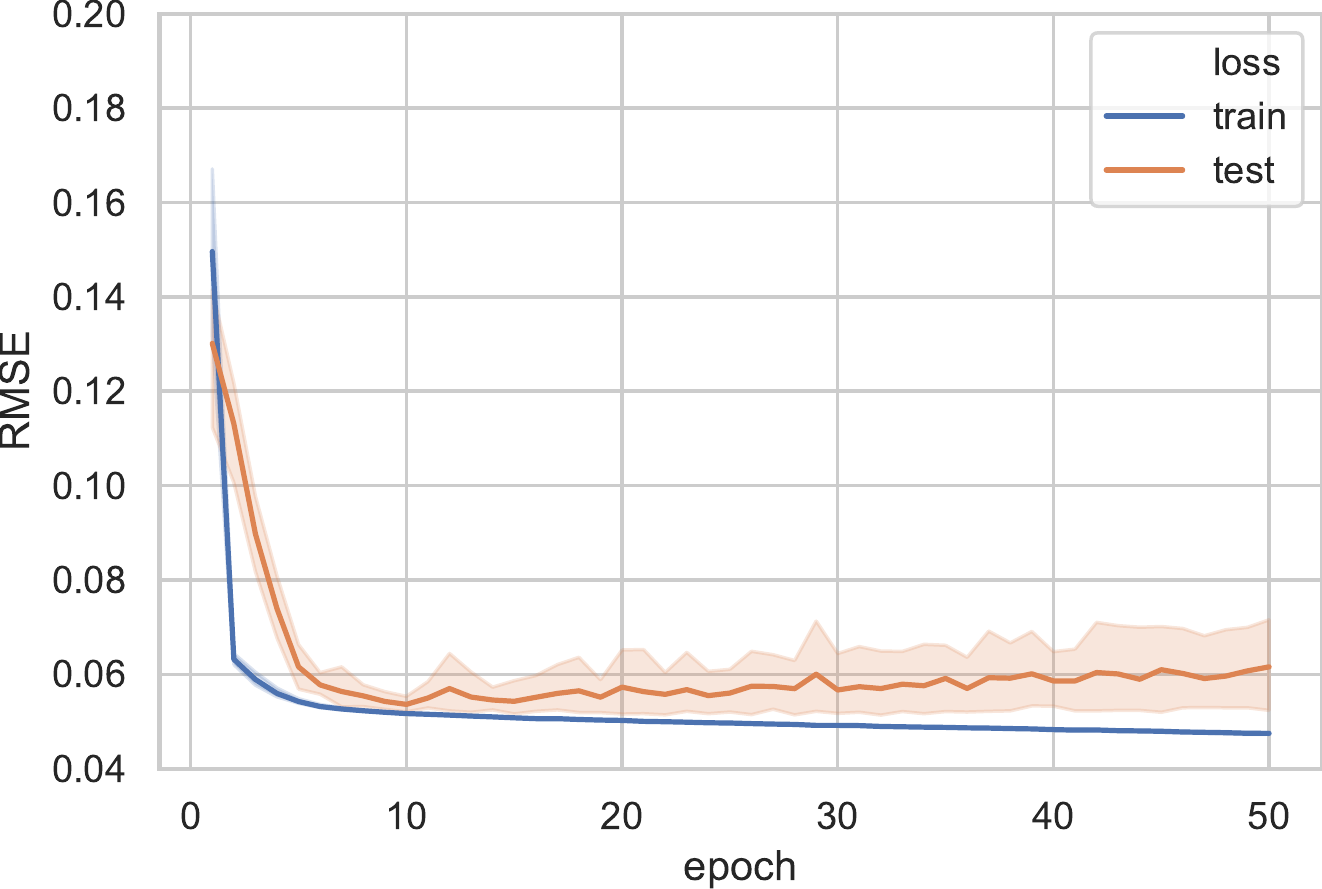}
     \end{subfigure}
     \begin{subfigure}[b]{0.5\textwidth}
         \includegraphics[width=0.95\textwidth]{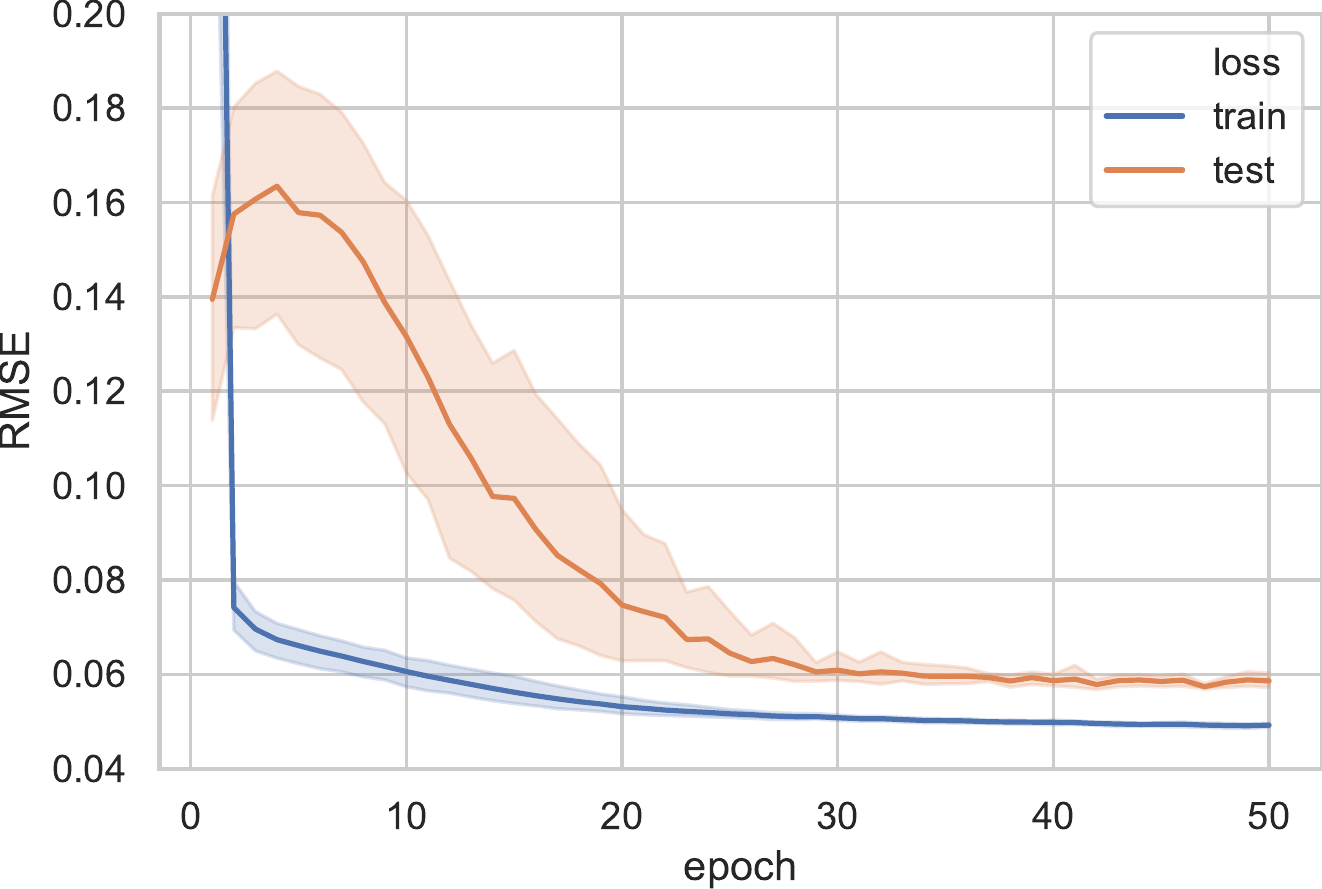}
     \end{subfigure}
     \caption{Progress of loss and test error over 50 epochs regarding zero shot prediction with two distinct splits. One training set consists of all graphs of length ${2, 3, 4, 5, 7}$ with a test set of the graphs of length $6$ (left). The other consists of all graphs of length ${2, 3, 4, 5, 6}$ with a test set of the graphs of length $7$ (right).}\label{fig:extrapolation}
\end{figure}

\subsubsection{Zero shot Prediction.} The progress of the training loss and test error for the \textit{zero shot prediction} case of our encoder can be seen in Figure \ref{fig:extrapolation}. The training set containing all graphs of length ${2, 3, 4, 5, 7} / {2, 3, 4, 5, 6}$ has a total amount of $361,614 / 64,542$ samples. Thus the encoder is tested only on  graphs of length $6 / 7$, which corresponds to a total of $62,010 / 359,082$ neural architectures. 
The experiments show that our model is able to accurately predict data that it has never seen before. The behaviour of the test error during the second zero shot prediction task, see Figure \ref{fig:extrapolation} (right), displays interesting information. During the first epochs, the error rises before it starts decreasing and approaching the training loss asymptotically. One interpretation could be that the model first learns simple graph properties like the number of nodes before it learns more complex graph substructures that generalise to the unseen data.

\subsection{Comparison to State of the Art}
In this section we compare our GNN-surrogate model with the most recent state-of-the-art predictor \cite{semisupervised}. They evaluate their predictor on the test accuracy of the NAS-Bench-101 dataset. Since predicting on the validation accuracy does not imply the same proper prediction behaviour on the test set, we evaluate our surrogate model in the same setting. In~\cite{semisupervised}, an auto-encoder model is first trained on the entire NAS-Bench-101 dataset and then fine-tuned with a graph similarity metric and test accuracy labels. Because the training relies on an unsupervised pre-training, they refer to the approach as semi-supervised.
To enable a direct comparison, we sample randomly $1,000/10,000/100,000$ graphs from the training data set and evaluate the performance prediction ability of the GNN surrogate model on all remaining $431,624/413,624/323,624$ graphs in the NAS-Bench-101 dataset. Please note that, at training time, the semi-supervised approach from \cite{semisupervised} actually has access to more data than our fully supervised approach, because of the unsupervised pre-training.

\begin{table}[t]
\caption{Comparison of predictive performance of surrogate models in terms of MSE on the test accuracies.}
\centering
\begin{tabular}{lcccc}
\toprule
Surrogate-Model                                              & \multicolumn{2}{c}{Performance Prediction}                               & \\
 & \footnotesize{$1,000$}
                                                            & \footnotesize{$10,000$}                     & \footnotesize{$100,000$}                   &  \\ \midrule
GNN Encoder    & $0.0044$       & $\mathbf{ 0.0022 }$   &$ \mathbf{0.0015}   $&   \\ 
Semi-Supervised Assessor \cite{semisupervised}\hspace{0.3cm}  & $\mathbf{ 0.0031 }$  & $0.0026   $& $0.0016 $&  \\

\bottomrule
\end{tabular}\label{tab:ass_rmse}
\end{table}

Table \ref{tab:ass_rmse} shows the experimental comparison, where we report an average over three runs for our approach while the numbers of \cite{semisupervised} are taken from their paper. The proposed GNN surrogate model surpasses the proposed semi-supervised assessor \cite{semisupervised} when 10.000 and 100.000 training architectures are available. With only 1.000 randomly drawn training samples, the results of our approach decrease. Yet, since we do not have access to the exact training samples used in \cite{semisupervised}, the results might become less comparable the lower the number of samples drawn.

    \section{Conclusion}\label{sec:conclusion}
    In this paper, we propose a GNN surrogate model for the prediction of the performance of neural architectures.  Through multiple experiments on NAS-Bench-101, we examined various capabilities of the encoder. The GNN encoder is a powerful tool regarding supervised  performance prediction and also especially in the zero-shot setup. Further research will mainly review the possibilities of neural architecture search in accordance with further performance prediction.
    
    \section*{Acknowledgement}
We thank Alexander Diete for helpful discussions and comments. This project is supported by the German Federal Ministry of Education and Research Foundation via the project DeToL.

	\bibliographystyle{splncs04}
	\bibliography{egbib}

\end{document}